\documentclass[11pt,a4paper]{article}
\usepackage[hyperref]{acl2020}
\usepackage{times}
\usepackage{latexsym}

\usepackage{microtype}

\usepackage{amsmath,amsfonts,bm}

\def\eqref#1{equation~\ref{#1}}

\def\1{\bm{1}}

\DeclareMathAlphabet{\mathsfit}{\encodingdefault}{\sfdefault}{m}{sl}
\SetMathAlphabet{\mathsfit}{bold}{\encodingdefault}{\sfdefault}{bx}{n}

\usepackage{booktabs} 
\usepackage{soul}
\usepackage{listings}
\usepackage{color}
\definecolor{lightgray}{gray}{0.9}
\usepackage{multirow}
\usepackage{amsmath}
\usepackage{amssymb}
\usepackage{graphicx}
\usepackage{caption}
\usepackage{float}
\usepackage{url}
\usepackage{xspace}
\usepackage{mathrsfs}
\usepackage{bm}
\usepackage{stfloats}
\usepackage{svg}
\usepackage{array}
\usepackage{enumitem}
\usepackage[export]{adjustbox}
\usepackage{amsthm}
\usepackage{algorithm}
\usepackage[noend]{algpseudocode}
\usepackage{wrapfig}

\algnewcommand\algorithmicinput{\textbf{INPUT:}}
\algnewcommand\INPUT{\item[\algorithmicinput]}

\algnewcommand\algorithmicoutput{\textbf{OUTPUT:}}
\algnewcommand\OUTPUT{\item[\algorithmicoutput]}
\aclfinalcopy 

\title{Rationalizing Medical Relation Prediction from Corpus-level Statistics}

\author{Zhen Wang$^1$, Jennifer Lee$^{2,3}$, Simon Lin$^{4}$, Huan Sun$^1$ \\
  $^1$The Ohio State University \\
  {$^2$Department of Family Medicine, The Ohio State University Wexner Medical Center} \\
  {$^3$Department of Physician Informatics, Nationwide Children’s Hospital} \\
  $^4$Abigail Wexner Research Institute at Nationwide Children's Hospital \\
  \texttt{\{wang.9215, sun.397\}@osu.edu} \\
  \texttt{\{Jennifer.Lee2, Simon.Lin\}@nationwidechildrens.org}
  }

\date{}

\begin{document}
\maketitle

\begin{abstract}
Nowadays, the interpretability of machine learning models is becoming increasingly important, especially in the medical domain. Aiming to shed some light on how to rationalize medical relation prediction, we present a new {interpretable} framework inspired by existing theories on how human memory works, e.g., theories of recall and recognition. Given the corpus-level statistics, i.e., a global co-occurrence graph of a {clinical} text corpus, to predict the relations between two entities, we first \textit{recall} rich contexts {associated with the target entities}, and then \textit{recognize} relational interactions between these contexts to form model rationales, which will contribute to the final prediction. We conduct experiments on a real-world public clinical dataset and show that our framework can not only achieve competitive predictive performance against a comprehensive list of neural baseline models, but also present rationales to justify its prediction. We further collaborate with medical experts deeply to verify the usefulness of our model rationales for clinical decision making. Code and datasets are available online\footnote{\url{https://github.com/zhenwang9102/X-MedRELA}}.
\end{abstract}

\section{Introduction}
\label{sec:intro}

Predicting relations between entities from a text corpus is a crucial task in order to extract structured knowledge, which can empower a broad range of downstream tasks, e.g., question answering~\cite{Xu2016QuestionAO}, dialogue systems~\cite{lowe2015incorporating}, reasoning~\cite{McCallum2016ChainsOR}, etc. There has been a large amount of existing work focusing on predicting relations based on \textit{raw texts} (e.g., sentences, paragraphs) mentioning two entities~\cite{hendrickx2009semeval, zeng-etal-2014-relation, zhou2016attention, mintz2009distant, riedel2010modeling, lin2016neural, Verga2018SimultaneouslyST, Yao2019DocREDAL}.

In this paper, we study a relatively new setting in which we predict relations between entities based on the \textit{global co-occurrence statistics} aggregated from a text corpus, and focus on medical relations and clinical texts in Electronic Medical Records (EMRs). The corpus-level statistics present a \textit{holistic {graph} view} of all entities in the corpus, which will greatly facilitate the relation inference, and can better preserve patient privacy than raw or even de-identified textual content and are becoming a popular substitute for the latter in the research community {for studying} EMR data~\cite{finlayson2014building, wang2019surfcon}.

\begin{figure}[t!]
    \centering
    \includegraphics[width=1\linewidth]{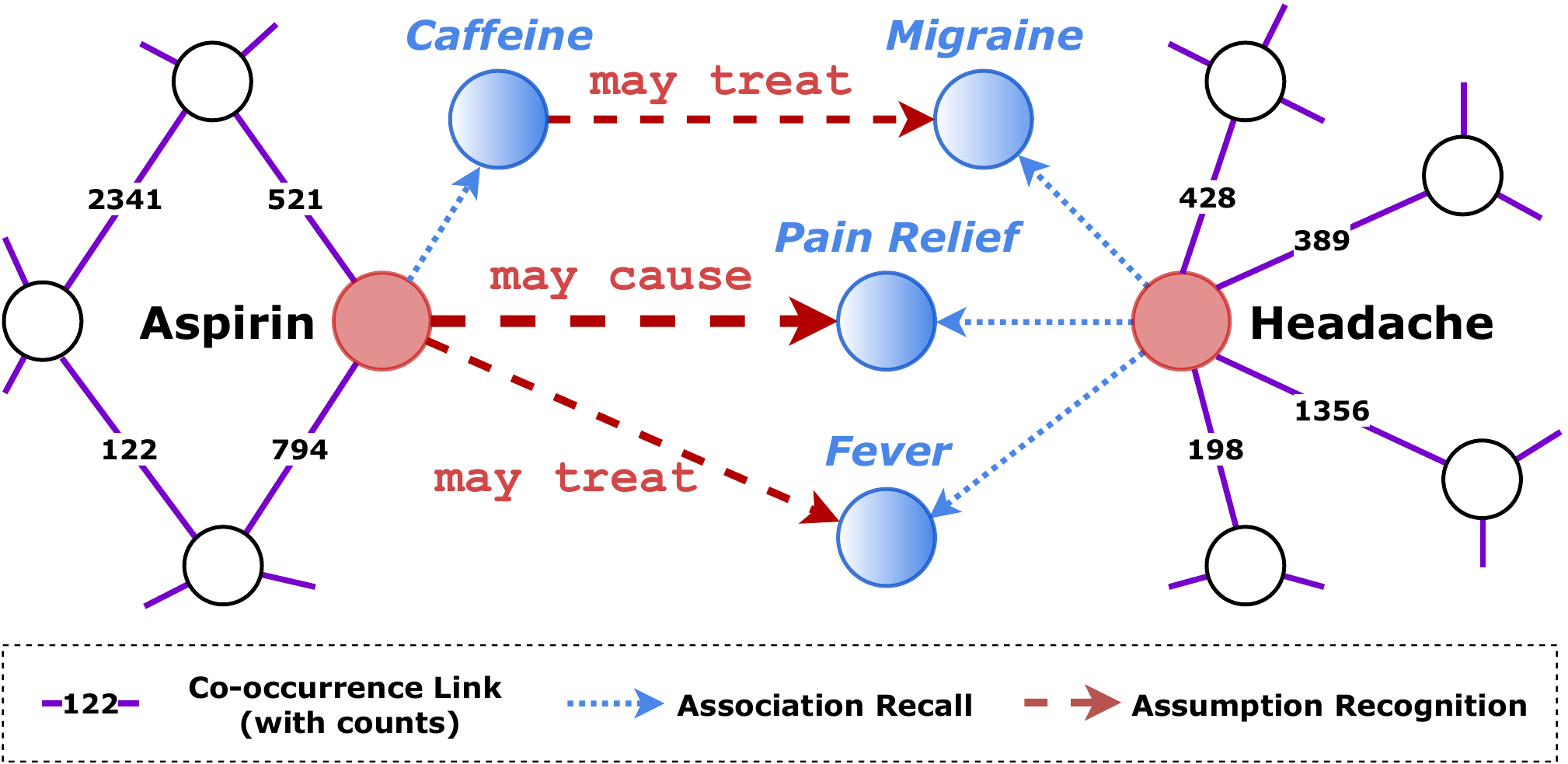}
    \caption{
    {Our intuition for how to rationalize relation prediction based on the {corpus-level} statistics. To infer the relation between the target entities (\textcolor{red}{red} nodes), we recall (\textcolor{blue}{blue} dashed line) their associated entities (\textcolor{blue}{blue} nodes) and infer their relational interactions (\textcolor{red}{red} dashed line), which will serve as assumptions or model rationales to support the target relation prediction.}
    }
    \label{fig:intuition_illustration}
    \vspace{-20pt}
\end{figure}

To predict relations between entities based on a global co-occurrence graph, intuitively, one can first {optimize the} graph embedding or global word embedding~\cite{pennington2014glove, perozzi2014deepwalk, tang2015line}, and then develop a relation classifier~\cite{nickel2011three, socher2013reasoning, yang2014embedding, wang2018multi} based on the embedding vectors of the two entities. However, such kind of neural frameworks often lack the desired \textit{interpretability}, which is especially important for the medical domain. In general, despite their superior predictive performance in many NLP tasks, the opaque decision-making process of neural models has concerned their adoption in high stakes domains like medicine, finance, and judiciary~\cite{rudin2019stop, murdoch2019definitions}. Building models that provide reasonable explanations and have increased transparency can remarkably enhance user trust~\cite{ribeiro2016should, miller2019explanation}. In  this  paper,  we  aim  to develop such a model for our medical relation prediction task.

To start with, we draw inspiration from the existing theories on cognitive processes about how human memory works, e.g., two types of memory retrieval (recall and recognition)~\cite{gillund1984retrieval}. Basically, in the \textit{recall} process, humans tend to retrieve contextual associations from long-term memory. For example, given the word ``Paris'', one may think of {``Eiffel Tower''} or {``France''}, which are strongly associated with ``Paris''~\cite{nobel2001retrieval, kahana2008associative, budiu2014memory}. Besides, there is a strong correlation between the association strength and the co-occurrence graph~\cite{spence1990lexical, lundberg2017unified}. In the \textit{recognition} process, {humans typically recognize if they have seen a certain piece of information before}. {Figure~\ref{fig:intuition_illustration} shows an example in the context of relation prediction.} Assume a model is to predict whether \textit{Aspirin} {may} treat \textit{Headache} or not (That ``\textit{Aspirin} may treat \textit{Headache}'' is a {known fact}, and we choose this relation triple for illustration purposes). {It is desirable if the model could perform the aforementioned two types of memory processes and produce rationales to base its prediction upon}: (1) Recall. What entities are associated with \textit{Aspirin}? What entities are associated with \textit{Headache}? (2) Recognition. Do those {associated} entities hold certain relations, which can be leveraged as clues to predict the target relation? {For instance, a model could first retrieve a relevant entity \textit{Pain Relief} for the tail entity \textit{Headache} {as they co-occur frequently}, and then recognize there is a chance that \textit{Aspirin} can lead to \textit{Pain Relief} (i.e., formulate model rationales or assumptions), based on which it could finally make a correct prediction (\textit{Aspirin}, {may treat}, \textit{Headache}).}

\begin{figure}
    \centering
    \includegraphics[width=1\linewidth]{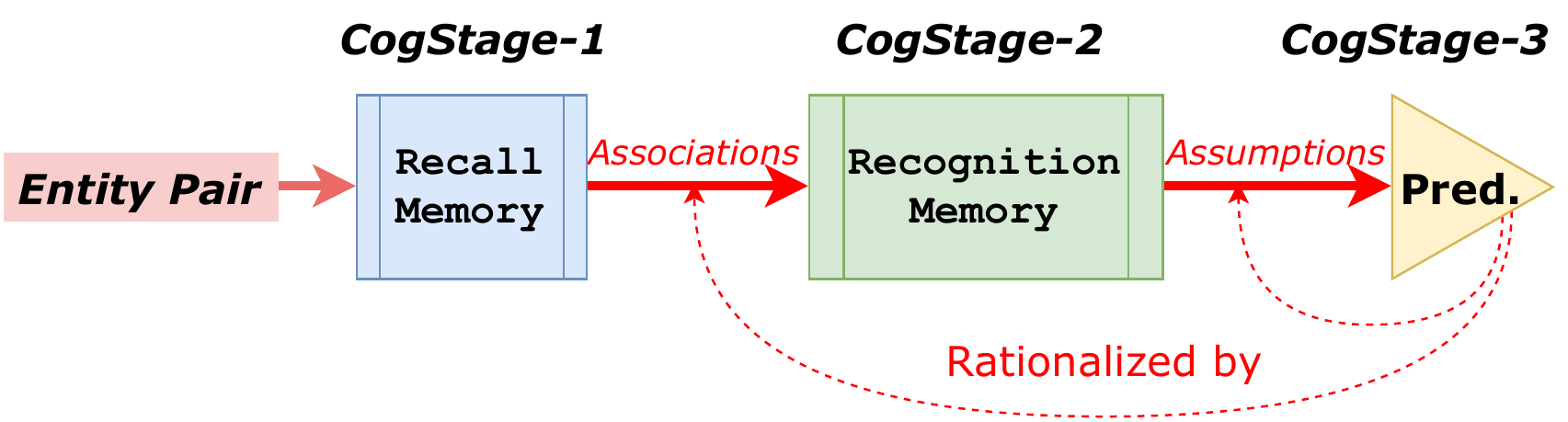}
    \vspace{-15pt}
    \caption{{A high-level illustration of our framework.}
    }
    \label{fig:workflow}
    \vspace{-15pt}
\end{figure}

{Now we formalize such intuition to rationalize the relation prediction task. Our framework consists of three stages, \textit{{global} association recall} (CogStage-1), \textit{assumption formation and representation} (CogStage-2), and \textit{prediction decision making} (CogStage-3), shown in Figure \ref{fig:workflow}. CogStage-1 models the process of recalling diverse contextual entities associated with the target head and tail entities respectively, CogStage-2 models the process of recognizing possible interactions between those recalled entities, which serve as model rationales (or, assumptions\footnote{We use the two terms interchangeably in this paper.}) and are represented as semantic vectors, and finally CogStage-3 aggregates all assumptions to infer the target relation.} {We jointly optimize all three stages using a training set of relation triples as well as the co-occurrence graph. Model rationales can be captured through this process \textit{without any gold rationales} available as direct supervision.} Overall, our framework rationalizes its relation prediction and is interpretable to users\footnote{Following \citet{murdoch2019definitions}, desired interpretability is supposed to provide insights to particular audiences, which in our case are medical experts.} by providing {justifications} for (i) why a particular prediction is made, (ii) how the assumptions of the prediction are developed, and (iii) how the particular assumptions are relied on.

On a real-life clinical text corpus, we compare our framework with various {competitive} methods to evaluate the predictive performance and interpretability. We show that our method obtains very competitive performance compared with a comprehensive list of various neural baseline models. Moreover, we follow recent work \cite{singh2018hierarchical, Jin2020Towards} to quantitatively evaluate model interpretability and demonstrate that rationales produced by our framework can greatly help earn expert trust. To summarize, we study the important problem of rationalizing medical relation prediction based on corpus-level statistics and propose a new framework inspired by cognitive theories, which outperforms competitive baselines in terms of both interpretability and predictive performance.
\vspace{-5pt}
\section{Background}
\vspace{-5pt}
\label{sec:pre}

\begin{figure*}[!t]
    \centering
    \includegraphics[width=1\textwidth]{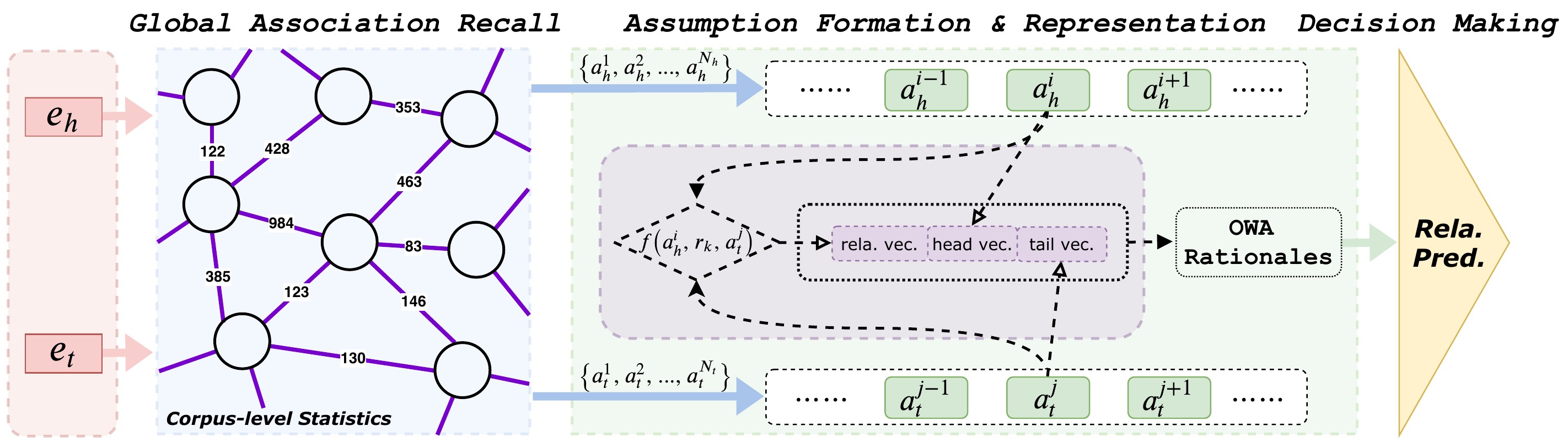}
    \vspace{-15pt}
    \caption{Framework Overview.}
    \label{fig:framewrok_overview}
    \vspace{-18pt}
\end{figure*}

{Different from existing work using raw texts for relation extraction, we assume a global co-occurrence graph (i.e., corpus-level statistics) is given, which was pre-constructed based on a text corpus $\mathcal{D}$, and denote it as an undirected graph $\mathcal{G}=(\mathcal{V}, \mathcal{E})$, where each vertex $v\in \mathcal{V}$ represents an entity extracted from the corpus and each edge $e\in \mathcal{E}$ is associated with the global co-occurrence count for the connected nodes. Counts reflect how frequent two entities appear in the same context (e.g., co-occur in the same sentence, document, or a certain time frame).} In this paper, we focus on clinical co-occurrence graph in which vertices are medical terms extracted from clinical notes. Nevertheless, as we will see later, our framework is very general and can be applied to other relations with corpus-level statistics.

Our motivation for working under this setting lies in three folds: (1) Such graph data is stripped of raw textual contexts and thus, has a better preserving of patient privacy~\cite{wang2019surfcon}, which makes itself easier to be constructed and shared under the HIPPA {protected environments}~\citep{act1996health} for medical institutes~\cite{finlayson2014building}; (2) Compared with open-domain relation extraction, entities holding a medical relation oftentimes do not co-occur in a local context (e.g., a sentence or paragraph). For instance, we observe that in a widely used clinical co-occurrence graph~\cite{finlayson2014building}, {which is also employed for our experiments later,} of all entity pairs holding the treatment relation according to UMLS (Unified Medical Language System), only about 11.4\% have a co-occurrence link (i.e., co-occur in clinical notes within a time frame like 1 day or 7 days); (3) {As suggested by cognitive theories~\cite{spence1990lexical}, lexical co-occurrence is significantly correlated with association strength in the recall memory process, which further inspires us to utilize such statistics to find associations and form model rationales for relation prediction.}

{Finally, our relation prediction task is formulated as: Given the global statistics $\mathcal{G}$ and an entity pair, we predict whether they hold a relation $r$ (e.g., \textsc{May\_treat}), and moreover provide a set of model rationales  $\mathcal{T}$ composed of relation triples for the prediction. For the example in Figure \ref{fig:intuition_illustration}, we aim to build a model that will not only accurately predict the \textsc{May\_treat} relation, but also provide meaningful rationales on how the prediction is made, which are crucial for gaining trust from clinicians.}

\vspace{-5pt}
\section{Methodology}
\vspace{-5pt}

Following a high-level framework illustration in Figure~\ref{fig:workflow}, we show a more detailed overview in Figure~\ref{fig:framewrok_overview} and introduce each component as follows.

\vspace{-5pt}
\subsection{CogStage-1: Global Association Recall}
\label{section:gcr}

{Existing cognitive theories~\cite{kahana2008associative} suggest that \textit{recall} is an essential function of human memory to retrieve \textit{associations} for later decision making. On the other hand, the association has been shown to significantly correlate with the lexical co-occurrence from the text corpus~\cite{spence1990lexical, lund1996producing}. Inspired by such theories and correlation, we explicitly build up our model based on recalled associations stemming from corpus-level statistics and provide global highly-associated contexts as the source of interpretations.}

{Given an entity,} we build an estimation module to globally infer associations based on the corpus-level statistics. Our module leverages distributional learning to fully explore the graph structure. {One can also directly utilize the raw neighborhoods in the co-occurrence graph, but due to the noise introduced in the preprocessing of building the graph, it is a less optimal choice in real practice.} 

Specifically, for a selected node/entity $e_i\in \mathcal{E}$, our global association recall module estimates a conditional probability $p\,(e_j|e_i)$, representing how likely the entity $e_j\in \mathcal{E}$ is associated with $e_i$\footnote{{We assume all existing entities can be possible associations for the given entity.}}. We formally define such conditional probability as:
\begin{equation}
\setlength{\abovedisplayskip}{5pt}
\setlength{\belowdisplayskip}{5pt}
\setlength{\abovedisplayshortskip}{0pt}
\setlength{\belowdisplayshortskip}{0pt}
    p\,(e_j|e_i)= 
    \frac{\exp \, ({\bm \upsilon'}_{e_j}^T\cdot \bm \upsilon_{e_i})}
    {\sum_{k=1}^{|\mathcal{V}|}\exp \, ({\bm \upsilon'}_{e_k}^T \cdot \bm \upsilon_{e_i})}
\end{equation}

\noindent where $\bm \upsilon_{e_i}\in \mathbb{R}^{d}$ is the embedding vector of node $e_i$ and $\bm \upsilon'_{e_j}\in \mathbb{R}^{d}$ is the context embedding for $e_j$. 

{There are many ways to approximate $p\,(e_j|e_i)$ from the global statistics, e.g., using global log-bilinear regression~\cite{pennington2014glove}. To estimate such probabilities and update entity embeddings efficiently,} we optimize the conditional distribution $p\,(e_j|e_i)$ to be close to the empirical distribution $\hat{p}\,(e_j|e_i)$ defined as:
\begin{equation}    
\setlength{\abovedisplayskip}{5pt}
\setlength{\belowdisplayskip}{5pt}
\setlength{\abovedisplayshortskip}{0pt}
\setlength{\belowdisplayshortskip}{0pt}
    \hat{p}\,(e_j|e_i)= \frac{p_{ij}}{\sum_{(i,k)\in \mathcal{E}} p_{ik}}
\end{equation}

\noindent where $\mathcal{E}$ is the set of edges in the co-occurrence graph and $p_{ij}$ is the PPMI value calculated by the co-occurrence counts between node $e_i$ and $e_j$. We adopt the cross entropy loss for the optimization:
\begin{equation}
\setlength{\abovedisplayskip}{5pt}
\setlength{\belowdisplayskip}{5pt}
\setlength{\abovedisplayshortskip}{0pt}
\setlength{\belowdisplayshortskip}{0pt}
\label{eqn:loss_context_pred}
    \mathcal{L}_{n}= -\sum_{(e_i,e_j) \in \mathcal{V}} \hat{p}(e_j|e_i)\ \text{log} \, (p(e_j|e_i))
\end{equation}

{This association recall module will be jointly trained with other objective functions to be introduced in the following sections. After that, given an entity $e_i$, we can select the top-$N_c$ entities from $p(\cdot |e_i)$ as $e_i$'s associative entities for subsequent assumption formation.}

\vspace{-5pt}
\subsection{CogStage-2: Assumption Formation and Representation}
\label{sec:CogStage2}

{As shown in Figure~\ref{fig:framewrok_overview}, with the associative entities from CogStage-1,} we are ready to \textit{formulate} and \textit{represent} assumptions. In this paper, we define model assumptions as \textit{relational interactions between associations}, that is, as shown in Figure~\ref{fig:intuition_illustration}, the model may identify (\textit{Caffeine}, \textsc{may\_treat}, \textit{Migraine}) as an assumption, which could help predict \textit{Aspirin} may treat \textit{Headache} (\textit{Caffeine} and \textit{Migraine} are associations for \textit{Aspirin} and \textit{Headache} respectively). {Such relational rationales are more concrete and much easier for humans to understand than the {widely-adopted explanation strategy \cite{yang2016hierarchical, Mullenbach2018ExplainablePO, vashishth2019attention} in NLP that is based on pure attention weights on local contexts}.}

{One straightway way to obtain such rationales is to query existing medical knowledge bases (KBs), e.g., (\textit{Caffeine}, \textsc{may\_treat}, \textit{Migraine}) may exist in SNOMED CT\footnote{\url{https://www.snomed.org/}} and can serve as a model rationale. We refer to rationales acquired in this way as the \textit{Closed-World Assumption} (CWA)~\cite{reiter1981closed} setting since only KB-stored facts are considered and trusted in a closed world. In contrast to the CWA rationales, considering the sparsity and incompleteness issues of KBs that are even more severe in the medical domain, we also propose the \textit{Open-World Assumptions} (OWA)~\cite{ceylan2016open} setting to discover richer rationales by estimating all potential relations between associative entities based on a seed set of relation triples (which can be regarded as prior knowledge).}

{In general, the CWA rationales are relatively more accurate {as each fact triple has been verified by the KB}, but would have a low coverage of other possibly relevant rationales for the target prediction. On the other hand, the OWA rationales are more comprehensive but could be noisy and less accurate, due to {the probabilistic estimation procedure} and the limited amount of prior knowledge. However, as we will see, by aggregating all OWA rationales into the whole framework {with an attention-based mechanism}, we can {select high-quality and most relevant rationales for prediction}. For the rest of the paper, by default we adopt the OWA setting in our framework and describe its details as follows.}

Specifically, given a pair of {head and tail} entity, $e_h, e_t\in \mathcal{V}$, let us denote their association sets as $\mathcal{A}(e_h)=\{a_h^i\}_{i=1}^{N_h}$ and $\mathcal{A}(e_t)=\{a_t^j\}_{j=1}^{N_t}$, where $N_h, N_t$ are the number of associative entities $a_h, a_t$ to use. Each entity has been assigned an embedding vector by the previous association recall module.
We first measure the probability of relations holding for the pair. Given $a_h^i\in \mathcal{A}(e_h), a_t^j\in \mathcal{A}(e_t)$ and a relation $r_k\in \mathcal{R}$, we define a scoring function as ~\citet{bordes2013translating} to estimate triple quality:
\begin{equation}
\setlength{\abovedisplayskip}{5pt}
\setlength{\belowdisplayskip}{5pt}
\setlength{\abovedisplayshortskip}{0pt}
\setlength{\belowdisplayshortskip}{0pt}
\label{eqn:scoring_func}
s_{k}^{ij} = f(a_h^i, r_k, a_t^j) = - ||\bm \upsilon_{a_h^i} + \bm \xi_k - \bm \upsilon_{a_t^j}||_1
\end{equation}

\noindent where $\bm \upsilon_{a_h^i}$ and $\bm \upsilon_{a_t^j}$ are embedding vectors, relations are parameterized by a relation matrix $R\in \mathbb{R}^{N_r\times d}$ and $\bm \xi_k$ is its $k$-th row vector. Such a scoring function encourages larger value for correct triples. Additionally, in order to filter unreliable estimations, we define an \texttt{NA} relation to represent other trivial relations {or no relation} as the score, $s_{\texttt{NA}}^{ij} = f(a_h^i, \texttt{NA}, a_t^j)$, which can be seen as a dynamic threshold to produce reasonable rationales.

Now we \textit{formulate} OWA rationales by calculating the conditional probability of a relation given a pair of associations as follows (we save the superscript $ij$ for space):
\begin{equation}
\label{eqn:threshold_prob}
\setlength{\abovedisplayskip}{5pt}
\setlength{\belowdisplayskip}{5pt}
\setlength{\abovedisplayshortskip}{0pt}
\setlength{\belowdisplayshortskip}{0pt}
p(r_k|a_h^i, a_t^j) =
 \begin{cases} 
  \dfrac{\exp \left(s_{k}\right)}{\sum_{s_{k} \geq s_{\texttt{NA}}} \exp \left(s_{k}\right)}, & s_{k} > s_{\texttt{NA}} \\
  0,       & s_{k}\leq s_{\texttt{NA}}
  \end{cases}
\end{equation}

{For each association pair, $(a_h^i, a_t^j)$, we only form an assumption with a relation $r_k^{*}$ if $r_k^{*}$ is top ranked according to $p(r_k|a_h^i, a_t^j)$\footnote{We remove the target relation to predict if it exists in the assumption set.}.}

To \textit{represent} assumptions, we integrate all relation information per pair into a single vector representation. Concretely, we calculate the assumption representation by treating $p(r_k|a_h^i, a_t^j)$ as weights for all relations as follows:
\begin{equation}
\label{eqn:rela_rep}
\setlength{\abovedisplayskip}{5pt}
\setlength{\belowdisplayskip}{5pt}
\setlength{\abovedisplayshortskip}{0pt}
\setlength{\belowdisplayshortskip}{0pt}
\mathsf{a}_{ij} = \rho(a_h^i, a_t^j;\mathcal{R}) = \sum_{k'=1}^{N_r} p(r_{k'}|a_h^i, a_t^j)\cdot \bm \xi_{k'}
\end{equation}

Finally, we combine the entity vectors as well as the relation vector to get the final representation of assumptions for association pair $(a_h^i, a_t^j)$, where $c_i\in \mathcal{A}(e_h)$ and $c_j\in \mathcal{A}(e_t)$:
\begin{equation}
\label{eqn:pair_rep}
\setlength{\abovedisplayskip}{5pt}
\setlength{\belowdisplayskip}{5pt}
\setlength{\abovedisplayshortskip}{0pt}
\setlength{\belowdisplayshortskip}{0pt}
\mathsf{e}_{ij} = \text{tanh}([\bm \upsilon_{a_h^i};\bm \upsilon_{a_t^j}; \mathsf{a}_{ij}]\bm W_p + \bm b_p)
\end{equation}

\noindent where $[\cdot\ ; \cdot]$ represents vector concatenation, $\bm W_p \in \mathbb{R}^{3d\times d_p}, \bm b_p \in \mathbb{R}^{d_p}$ are the weight matrix and bias in a fully-connected network.

\vspace{-5pt}
\subsection{CogStage-3: Prediction Decision Making}
\label{sec:ril}

Analogical to human thinking, our decision making module aggregates all assumption representations and measures their accountability for the final prediction. It learns a distribution over all assumptions and we select the ones with highest probabilities as model rationales. More specifically, we define a scoring function $g(\mathsf{e}_{ij})$ to estimate the accountability based on the assumption representation $\mathsf{e}_{ij}$ and normalize $g(\mathsf{e}_{ij})$ as:
\begin{equation}
\setlength{\abovedisplayskip}{5pt}
\setlength{\belowdisplayskip}{5pt}
\setlength{\abovedisplayshortskip}{0pt}
\setlength{\belowdisplayshortskip}{0pt}
\label{eqn:att_weight}
g(\mathsf{e}_{ij}) = \bm v^T \cdot \text{tanh}(\bm W_a \mathsf{e}_{ij}+\bm b_a)
\end{equation}
\begin{equation}
\setlength{\abovedisplayskip}{5pt}
\setlength{\belowdisplayskip}{5pt}
\setlength{\abovedisplayshortskip}{0pt}
\setlength{\belowdisplayshortskip}{5pt}
\label{eqn:plausibility}
p_{ij} = \frac{\exp ({g(\mathsf{e}_{ij})})}{\sum_{m=1}^{N_h} \sum_{n=1}^{N_t} \exp (g(\mathsf{e}_{mn}))}
\end{equation}

\noindent where $\bm W_a, \bm b_a$ are the weight matrix and bias for the scoring function. Then we get the weighted rationale representation as:
\begin{equation}
\setlength{\abovedisplayskip}{5pt}
\setlength{\belowdisplayskip}{5pt}
\setlength{\abovedisplayshortskip}{0pt}
\setlength{\belowdisplayshortskip}{0pt}
\mathsf{r} = \psi(e_h, e_t)=\sum_{i=1}^{N_h} \sum_{j=1}^{N_t} p_{ij} \mathsf{e}_{ij}
\end{equation}

With the representation of weighted assumption information for the target pair $(e_h, e_t)$, we calculate the binary prediction probability for relation $r$ as:
\begin{equation}
\setlength{\abovedisplayskip}{5pt}
\setlength{\belowdisplayskip}{5pt}
\setlength{\abovedisplayshortskip}{0pt}
\setlength{\belowdisplayshortskip}{0pt}
\label{eqd:pred}
p(r|e_h, e_t) = \sigma (\bm W_r \mathsf{r} + \bm b_r)
\end{equation}

\noindent where $\sigma(x) = 1/(1 + \exp(-x))$ and $\bm W_r, \bm b_r$ are model parameters.

\noindent \textbf{Rationalizing relation prediction.} After fully training the entire model, to recover the most contributing assumptions for predicting the relation between the given target entities $(e_h, e_t)$, we compute the importance scores for all assumptions and select those most important ones as model rationales. In particular, we multiply $p_{ij}$ (the weight for association pair $(a_h^i, a_t^j)$ in Eqn.~\ref{eqn:plausibility}) with $p(r_k|a_h^i, a_t^j)$ (the probability of a relation given the pair $(a_h^i, a_t^j)$ in Eqn.~\ref{eqn:threshold_prob}) to score the triple $(a_h^i, r_k, a_t^j)$. We rank all such triples for $a_h^i \in \mathcal{A}(e_h), a_t^j \in \mathcal{A}(e_t), r_k\in \mathcal{R}$ and select the top-$K$ triples as model rationales for the final relation prediction.

\vspace{-5pt}
\subsection{Training}
\label{sec:training}

We now describe how we train our model efficiently for multiple modules. For relational learning to estimate the conditional probability $p(r_k|a_h^i, a_t^j)$, we utilize training data as the seed set of triples for all relations as correct triples denoted as $(h, r, t) \in \mathcal{P}$. The scoring function in Eqn.~\ref{eqn:scoring_func} is expected to score higher for correct triples than the corrupted ones in which we denote $\mathcal{N}(?, r, t)$ ($\mathcal{N}(t, r, ?)$) as the set of corrupted triples by replacing the head (tail) entity randomly. Instead of using margin-based loss function, we adopt a more efficient training strategy from \citep{kadlec2017knowledge, toutanova2015observed} with a negative log likelihood loss function as:
\begin{equation}
\setlength{\abovedisplayskip}{5pt}
\setlength{\belowdisplayskip}{5pt}
\setlength{\abovedisplayshortskip}{0pt}
\setlength{\belowdisplayshortskip}{0pt}
\label{eqn:loss_nllloss_rela}
\begin{split}
\mathcal{L}_r = &- \textstyle \sum_{\left(h, r, t\right) \in \mathcal{P}} \log p\left(h | t, r\right)  \\
 & -\textstyle  \sum_{\left(h, r, t\right) \in \mathcal{P}} \log p\left(t | h, r\right)
\end{split}
\end{equation}

\noindent where the conditional probability $p(h|t, r)$ is defined as follows ($p(t|h, r)$ is defined similarly): 
\begin{equation}
\setlength{\abovedisplayskip}{5pt}
\setlength{\belowdisplayskip}{5pt}
\setlength{\abovedisplayshortskip}{0pt}
\setlength{\belowdisplayshortskip}{0pt}
\label{eqn:rela_sample_prob}
p\left(h | t, r \right)=\frac{\exp ({f\left(h, r, t \right)}) }{\sum_{h' \in \mathcal{N}\left(?, r, t\right)} \exp ({f\left(h', r, t\right)})}
\end{equation}

For our binary relation prediction task, we define a binary cross entropy loss function with Eqn.~\ref{eqd:pred} as follows:
\begin{equation}
\setlength{\abovedisplayskip}{5pt}
\setlength{\belowdisplayskip}{5pt}
\setlength{\abovedisplayshortskip}{0pt}
\setlength{\belowdisplayshortskip}{0pt}
\label{eqn:loss_bce}
\begin{split}
\mathcal{L}_p =  - &\textstyle \sum_{i=1}^{M} (y_i\cdot \text{log} (p(r|e_h^i, e_t^i)) \\
&+ (1 - y_i)\cdot \text{log}(1 - p(r|e_h^i, e_t^i)))
\end{split}
\end{equation}

\noindent where $M$ is the number of samples, $y_i$ is the label {{showing whether $e_h, e_t$ holds a certain relation}}.

The above three loss functions, i.e., $\mathcal{L}_n$ for global association recall, $\mathcal{L}_r$ for relational learning and $\mathcal{L}_p$ for relation prediction, are all jointly optimized. All three of them share the entity embeddings and $\mathcal{L}_p$ will reuse the relation matrix from $\mathcal{L}_r$ to conduct the rationale generation. 
Please see appendix for more details of our training algorithm.

\vspace{-5pt}
\section{Experiments}
\vspace{-5pt}

In this section, we first introduce our experimental setup, e.g, the corpus-level co-occurrence statistics and datasets used for our experiments, and then compare our model with a list of comprehensive competitive baselines in terms of predictive performance. Moreover, we conduct expert evaluations as well as case studies to demonstrate the usefulness of our model rationales.

\vspace{-5pt}
\subsection{Dataset}

We directly adopt a publicly available medical co-occurrence graph for our experiments~\citep{finlayson2014building}. The graph was constructed in the following way: \citet{finlayson2014building} first used an efficient annotation tool~\cite{lependu2012annotation} to extract medical terms from 20 million clinical notes collected by Stanford Hospitals and Clinics, and then computed the co-occurrence counts of two terms based on their appearances in one patient's records within a certain time frame (e.g., 1 day, 7 days). We experiment with their biggest dataset with the largest number of nodes (i.e., the per-bin 1-day graph here\footnote{\url{https://datadryad.org/stash/dataset/doi:10.5061/dryad.jp917}}) {so as to have sufficient training data.} The co-occurrence graph contains 52,804 nodes and 16,197,319 edges.

To obtain training labels for relation prediction, we utilize the mapping between medical terms and concepts provided by \citet{finlayson2014building}. To be specific, they mapped extracted terms to UMLS concepts with a high mapping accuracy by suppressing the least possible meanings of each term (see \citet{finlayson2014building} for more details). We utilize such mappings to automatically collect relation labels from UMLS. For term $e_a$ and $e_b$ that are respectively mapped to medical concept $c_A$ and $c_B$, we find the relation between $c_A$ and $c_B$ in UMLS, which will be used as the label for $e_a$ and $e_b$.

{Following \citet{wang2014medical} that studied distant supervision in medical text and identified several crucial relations for clinical decision making, we select 5 important medical relations with no less than 1,000 relation triples in our dataset.} {Each relation is mapped to UMLS semantic relations, e.g., relation \textsc{Causes} corresponds to \textit{cause\_of}, \textit{induces}, \textit{causative\_agent\_of} in UMLS. A full list of mapping is in the appendix.} We sample an equal number of negative pairs by randomly pairing head and tail entities with the correct argument types~\cite{wang2016building}. We split all samples into train/dev/test sets with a ratio of 70/15/15. Only relation triples in the training set are used to optimize relational parameters. The statistics of the positive samples for relations are summarized in Table~\ref{tab:cooc-statistics}. 

\begin{table}[t!]
\centering
\resizebox{0.9\linewidth}{!}{
\begin{tabular}{@{}lccc@{}}
\toprule
Med Relations   & Train                      & Dev                       & Test                      \\ \midrule
Symptom of     & 14,326                     & 3,001                     & 3,087                     \\
May treat      & 12,924                     & 2,664                     & 2,735                     \\
Contraindicates & 10,593                     & 2,237                     & 2,197                     \\
May prevent    & 2,113                      & 440                       & 460                       \\
Causes          & 1,389                      & 305                       & 354                       \\ \midrule
Total           & 41.3k                        & 8.6k                       & 8.8k                       \\ \bottomrule
\end{tabular}}
\vspace{-5pt}
\caption{Dataset Statistics.}
\label{tab:cooc-statistics}
\vspace{-15pt}
\end{table}

\begin{table*}[]
\centering
\resizebox{1\linewidth}{!}{
\begin{tabular}{@{}lcccccc@{}}
\toprule
Methods & \textsc{May\_treat} & \textsc{Contrain.} & \textsc{Symptom\_of} & \textsc{May\_prevent} & \textsc{Causes} & Avg. \\ \midrule
Word2vec + DistMult & 0.767 ($\pm$0.008) & 0.777 ($\pm$0.013) & 0.815 ($\pm$0.005) & 0.649 ($\pm$0.018) & 0.671 ($\pm$0.015) & 0.736 \\
Word2vec + RESCAL & 0.743 ($\pm$0.010) & 0.767 ($\pm$0.003) & 0.808 ($\pm$0.009) & 0.658 ($\pm$0.023) & 0.659 ($\pm$0.039) & 0.727 \\
Word2vec + NTN & 0.693 ($\pm$0.013) & 0.758 ($\pm$0.005) & 0.808 ($\pm$0.004) & 0.605 ($\pm$0.022) & 0.631 ($\pm$0.017) & 0.699 \\ \midrule
DeepWalk + DistMult & 0.740 ($\pm$0.003) & 0.776 ($\pm$0.004) & 0.805 ($\pm$0.003) & 0.608 ($\pm$0.014) & 0.650 ($\pm$0.018) & 0.716 \\
DeepWalk + RESCAL & 0.671 ($\pm$0.010) & 0.778 ($\pm$0.003) & 0.800 ($\pm$0.003) & 0.600 ($\pm$0.023) & \textbf{0.708 ($\pm$0.011)} & 0.711 \\
DeepWalk + NTN & 0.696 ($\pm$0.006) & 0.778 ($\pm$0.005) & 0.787 ($\pm$0.005) & 0.614 ($\pm$0.016) & 0.674 ($\pm$0.024) & 0.710 \\
LINE + DistMult & 0.767 ($\pm$0.003) & 0.783 ($\pm$0.002) & 0.795 ($\pm$0.003) & 0.621 ($\pm$0.015) & 0.641 ($\pm$0.024) & 0.721 \\
LINE + RESCAL & 0.725 ($\pm$0.003) & 0.771 ($\pm$0.002) & 0.801 ($\pm$0.001) & 0.613 ($\pm$0.013) & 0.694 ($\pm$0.015) & 0.721 \\
LINE + NTN & 0.733 ($\pm$0.002) & 0.773 ($\pm$0.003) & 0.800 ($\pm$0.001) & 0.601 ($\pm$0.015) & 0.706 ($\pm$0.013) & 0.723 \\ \midrule
REPEL-D + DistMult & 0.784 ($\pm$0.002) & 0.797 ($\pm$0.002) & 0.809 ($\pm$0.003) & 0.681 ($\pm$0.010) & 0.694 ($\pm$0.022) & 0.751 \\
REPEL-D + RESCAL & 0.726 ($\pm$0.003) & 0.780 ($\pm$0.002) & 0.776 ($\pm$0.002) & \textbf{0.685 ($\pm$0.010)} & \textbf{0.708 ($\pm$0.003)} & 0.737 \\
REPEL-D + NTN & 0.736 ($\pm$0.004) & 0.780 ($\pm$0.002) & 0.773 ($\pm$0.001) & 0.667 ($\pm$0.015) & 0.694 ($\pm$0.024) & 0.731 \\ \midrule
Ours (w/ CWA) & 0.709 ($\pm$0.005) & 0.751 ($\pm$0.009) & 0.744 ($\pm$0.007) & 0.667 ($\pm$0.008) & 0.661 ($\pm$0.032) & 0.706 \\
Ours & \textbf{0.805 ($\pm$0.017)} & \textbf{0.811 ($\pm$0.006)} & \textbf{0.816 ($\pm$0.004)} & 0.676 ($\pm$0.020) & 0.684 ($\pm$0.017) & \textbf{0.758} \\ \bottomrule
\end{tabular}
}
\vspace{-5pt}
\caption{Comparison of model predictive performance. We run all methods for five times and report the averaged F1 scores with standard deviations.}
\label{tab:main_results}
\vspace{-12pt}
\end{table*}

\vspace{-5pt}
\subsection{Predictive Performance Evaluation}

\noindent \textbf{Compared Methods.} There are a number of advanced neural methods {\cite{tang2015line, qu2018weakly, wang2018multi} that have been developed for the link prediction task}, i.e.,  predicting the relation between two nodes in a co-occurrence graph. {At the high level, their frameworks comprise of an entity encoder and a relation scoring function.} We adapt various existing methods for both the encoder and {the scoring functions} for comprehensive comparison.
Specifically, given the co-occurrence graph, we employ existing distributional representation learning methods to learn entity embeddings. With the entity embeddings as input features, we adapt various models from the knowledge base completion literature as a binary relation classifier. More specifically, for the encoder, we select one word embedding method, Word2vec~\cite{mikolov2013distributed, levy2014neural}, two graph embedding methods, random-walk based DeepWalk~\cite{perozzi2014deepwalk}, edge-sampling based LINE~\cite{tang2015line}, and one distributional approach {REPEL-D~\cite{qu2018weakly}} for weakly-supervised relation extraction that leverages both the co-occurrence graph and training relation {triples} to learn entity representations. For the {scoring functions}, we choose DistMult~\cite{yang2014embedding}, RESCAL~\cite{nickel2011three} and NTN~\cite{socher2013reasoning}. 

{Note that one can apply more complex encoders or scoring functions to obtain higher predictive performance; however, {in this work, we emphasize more on model interpretability than predictive performance, and unfortunately,} all such frameworks are hard to interpret as they provide little or no explanations on how predictions are made.}

{We also show the predictive performance of our framework under the CWA setting in which the CWA rationales are existing triples in a ``closed'' knowledge base {(i.e., UMLS)}. We first adopt the pre-trained association recall module to retrieve associative contexts for head and tail entities, {then formulate the assumptions using top-ranked triples (that exist in our relation training data), where the rank is based on the product of their retrieval probabilities ($p_{ij}=p(e_i|e_h)\times p(e_j|e_t)$). We keep the rest of our model the same as the OWA setting.}}

\vspace{5pt}
\noindent \textbf{Results.} 
We compare the predictive performance of different models in terms of F1 score under each relation prediction task. As shown in Table~\ref{tab:main_results}, our model obtains very competitive performance compared with a comprehensive list of baseline methods. Specifically, on the prediction tasks of \textsc{May\_treat} and \textsc{Contraindicates}, our model achieves a substantial improvement (1$\sim$2 F1 score) and a very competitive performance on the task of \textsc{Symptom\_of} and \textsc{May\_prevent}. The small amount of training data might partly explain why our model does not perform so well in the \textsc{Causes} task. Such comparison shows the {effectiveness} of predicting relations based on associations and their relational interactions. Moreover, compared with those baseline models which encode graph structure into latent vector representation, our model utilizes co-occurrence graph {more explicitly} by leveraging the associative contexts {symbolically} to generate human-understandable rationales, which can assist medical experts as we will see shortly. {In addition, we observe that our model consistently outperforms the CWA setting: {Despite the CWA rationales are true statements on their own, they tend to have a low coverage of possible rationales, and thus, may be not so relevant for the target relation prediction, which leads to a poor predictive performance.}}

\vspace{-5pt}
\subsection{Model Rationale Evaluation}
\label{sec:ration_eval}

To measure the quality of our model rationales (i.e., OWA rationales), as well as to conduct an ablation study of our model, we conduct an expert evaluation for the OWA rationales and also compare them with the CWA rationales. {We first collaborate with {a physician} to explore how much a model's rationales help them better trust the model's prediction following recent work for evaluating model interpretability~\cite{singh2018hierarchical, Mullenbach2018ExplainablePO, atutxa2019interpretable, Jin2020Towards}. Then, we present some case studies to show what kind of rationales our model has learnt.} Note that compared with evaluation by human annotators for open-domain tasks (without expertise requirement), evaluation by medical experts is more challenging in general. The physician in our study (an M.D. with 9 years of clinical experience and currently a fellow trained in clinical informatics), who is able to understand the context of terms and the basics of the compared algorithms and can dedicate time, is qualified for our evaluation. 

\begin{table}[t!]
\centering
\resizebox{1\linewidth}{!}{
\begin{tabular}{@{}lcc@{}}
\toprule
 & OWA Rationales & CWA Rationales  \\ \midrule
Ranking Score & 17 & 5  \\
Avg. Sum Score/Case & 6.14 & 2.24  \\
Avg. Max Score/Case & 2.04 & 0.77  \\ \bottomrule
\end{tabular}
}
\vspace{-5pt}
\caption{Human evaluation on the quality of rationales.}
\label{tab:human_eval}
\vspace{-18pt}
\end{table}

\vspace{5pt}
\noindent \textbf{Expert Evaluation.} 
We first explained to the physician about the recall and recognition process in our framework and how model rationales are developed. They endorsed such reasoning process as one possible way to gain their trust in the model. 
Next, for each target pair for which our model correctly makes the prediction, they were shown the top-5 rationales produced by our framework and were asked whether each rationale helps them better trust the model prediction. For each rationale, they were asked to score it from 0 to 3 in which 0 is \textit{no helpful}, 1 is \textit{a little helpful}, 2 is \textit{helpful} and 3 is \textit{very helpful}. In addition to the individual rationale evaluation, we further compare the overall quality of CWA and OWA rationales, by letting experts rank them based the helpfulness of each set of rationales (the rationale set ranked higher gets 1 ranking score and {both get 0 if they have the same rank}). We refer readers to the appendix for more details of the evaluation protocol. We randomly select 30 cases in the \textsc{May\_treat} relation and the overall evaluation results are summarized in Table~\ref{tab:human_eval}. Out of 30, OWA wins in 17 cases and gets higher scores {on individual rationales} per case on average. {There are 8 cases where the two sets of rationales are ranked the same\footnote{{Of which, 7 cases are indicated equally unhelpful}.} and 5 cases where CWA is better.} {To get a better idea of how the OWA model obtains more trust, we calculate the {average sum score per case}, which shows the OWA model gets a higher overall score per case. {Considering in some cases only a few rationales are able to get non-zero scores, we also calculate the average max score per case, which shows that our OWA model generally provides one \textit{helpful} rationale (score$>$2) per case}.}
Overall, as we can see, the OWA rationales {are more helpful to gain expert trust}.

\begin{table}[t!]
\centering
\resizebox{0.98\linewidth}{!}{
\begin{tabular}{@{}ccc@{}}
\toprule
\multicolumn{3}{c}{Case 1}                                                              \\ \midrule
cephalosporins   & may\_treat            & bacterial infection           \\ \midrule
cefuroxime      & may\_treat          & viral syndrome           \\
cefuroxime      & may\_treat          & low grade fever                    \\
\textbf{cefuroxime}            & \textbf{may\_treat}  & \textbf{infectious diseases}  \\
cefuroxime   & may\_prevent & low grade fever           \\
sulbactam        & may\_treat & low grade fever  \\ 
\midrule
\multicolumn{3}{c}{Case 2}                                                              \\ \midrule
azelastine               & may\_treat            & perennial allergic rhinitis          \\ \midrule
\textbf{astepro}         & \textbf{may\_treat}  & \textbf{perennial allergic rhinitis} \\
\textbf{pseudoephedrine} & \textbf{may\_treat}  & \textbf{perennial allergic rhinitis} \\
\textbf{ciclesonide}     & \textbf{may\_treat}  & \textbf{perennial allergic rhinitis} \\
overbite                 & may\_treat           & perennial allergic rhinitis          \\
diclofenac               & may\_treat           & perennial allergic rhinitis          \\ \bottomrule
\end{tabular}
}
\vspace{-5pt}
\caption{Case studies for rationalizing medical relation prediction. For each case, the first panel is target pair and the second is top-5 rationales (\textbf{Bold} ones are useful rationales {with high scores from the physician}). The left (right) most column is the head (tail) term and their {relational} associations.}
\label{tab:case_stuidies}
\vspace{-15pt}
\end{table}

\vspace{5pt}
\noindent \textbf{Case Study.}
{Table~\ref{tab:case_stuidies} shows two concrete examples demonstrating what kind of model rationales our framework bases its predictions on. {We highlight the rationales that receive high scores from the physician for being especially useful for trusting the prediction.} 
{As we can see, our framework is able to make correct predictions based on reasonable rationales. For instance, to predict that ``cephalosporine'' may treat ``bacterial infection'', our model relies on the rationale that ``cefuroxime'' may treat ``infectious diseases''.  
We also note that not all rationales are clinically established facts {or even make sense}, due to the unsupervised rationale learning and the probabilistic assumption formation process, which leaves space for future work to further improve the quality of rationales.} {Nevertheless,} such model rationales can provide valuable information or new insights for clinicians. For another example, as {pointed out} by the physician, {different medications possibly having the same treatment response, as shown in Case 2, could be clinically useful. {That is, if three medications are predicted to possibly treat the same condition and a physician is only aware of two doing so, one might get insights into trying the third one.}} To summarize, our model is able to provide reasonable rationales and help users understand how model predictions are made in general.}

\vspace{-5pt}
\section{Related Work}
\label{sec:related}
\vspace{-5pt}

{Relation Extraction (RE) typically focuses on predicting relations between two entities based on their text mentions, and has been well studied in both open domain~\cite{mintz2009distant, zeng2015distant, riedel2013relation, lin2016neural, song2019leveraging, deng-sun-2019-leveraging} and biomedical domain~\cite{uzuner20112010, wang2014medical, sahu-etal-2016-relation, lv2016clinical, he2019classifying}. Among them, most state-of-the-art work develops various powerful neural models by leveraging human annotations, linguistic patterns, distance supervision, etc. More recently, an increasing amount of work has been proposed to improve model's transparency and interpretability. For example, \citet{lee2019semantic} visualizes self-attention weights learned from BERT~\cite{Devlin2019BERTPO} to explain relation prediction. However, such text-based interpretable models tend to provide explanations within a \textit{local context} (e.g., {words in a single sentence mentioning target entities}), which may not capture a \textit{holistic view} of all entities and their relations stored in a text corpus. We believe that such a holistic view is important for interpreting relations and can be provided {to some degree} by the \textit{global statistics} from a text corpus. Moreover, global statistics have been widely used in the clinical domain as they can better preserve patient privacy~\cite{finlayson2014building, wang2019surfcon}.}

{On the other hand, in recent years, graph embedding techniques~\cite{perozzi2014deepwalk, tang2015line, node2vec-kdd2016, yue2020graph} have been widely applied to learn node representations based on graph structure. Representation learning based on global statistics from a text corpus (i.e., co-occurrence graph) has also been studied~\cite{, levy2014neural, pennington2014glove}. After employing such methods to learn entity embeddings, a number of relation classifiers~\cite{nickel2011three, bordes2013translating, socher2013reasoning, yang2014embedding, wang2018multi} can be adopted for relation prediction. {We compare our method with such frameworks to show its competitive predictive accuracy. However, such frameworks tend to be difficult to interpret as they provide little or no explanations on how decisions are made. In this paper, we focus more on model interpretability than predictive accuracy, and draw inspirations from existing cognitive theories of recall and recognition to develop a new framework, which is our core contribution.}}

{Another line of research related to interpreting relation prediction is path-based knowledge graph (KG) reasoning~\cite{gardner2014incorporating,  neelakantan2015compositional, Guu2015TraversingKG, Xiong2017DeepPathAR, stadelmaier2019modeling}. In particular, existing paths mined from millions of relational links {in knowledge graphs} can be used to provide justifications for relation predictions. {For example, to explain \textit{Microsoft} and \textit{USA} may hold the relation \textit{CountryOfHeadquarters}, by traversing a KG, one can extract the path \textit{Microsoft} $\xrightarrow{\text{IsBasedIn}}$ \textit{Seattle} $\xrightarrow{\text{CountryLocatedIn}}$ \textit{USA} as one explanation.} However, such path-finding methods typically require large-scale relational links to infer path patterns, and cannot be applied to our co-occurrence graph as the co-occurrence links are unlabeled.}

{In addition, our work is closely related to the area of rationalizing machine decision by generating justifications/rationales accounting for model's prediction.} {In some scenarios, human rationales are provided as extra supervision for more explainable models~\cite{zaidan2007using, Bao2018DerivingMA}. However, due to the high cost of manual annotation, model rationales are desired to be learned in an unsupervised manner\cite{Lei2016RationalizingNP, bouchacourt2019educe, zhao2019riker}. For example, \citet{Lei2016RationalizingNP} select a subset of words as rationales and \citet{bouchacourt2019educe} provide an explanation based on the absence or presence of ``concepts'', {where the selected words and ``concepts'' are learned unsupervisedly.} Different from text-based tasks, in this paper, we propose to rationalize relation prediction based on global co-occurrence statistics and similarly, model rationales in our work are captured without explicit manual annotation either, via a joint training framework.}

\vspace{-5pt}
\section{Conclusion}
\vspace{-5pt}

In this paper, we propose an interpretable framework to rationalize medical relation prediction based on corpus-level statistics. Our framework is inspired by existing cognitive theories on human memory recall and recognition, and can be easily understood by users as well as provide reasonable explanations to justify its prediction. {Essentially, it leverages corpus-level statistics to recall associative contexts and recognizes their relational connections as model rationales.} Compared with a comprehensive list of baseline models, our model obtains competitive predictive performances. Moreover, we demonstrate its interpretability via expert evaluation and case studies.

\section*{Acknowledgments}

We thank Srinivasan Parthasarathy, Ping Zhang, Samuel Yang and Kaushik Mani for valuable discussions. We also thank the anonymous reviewers for their hard work and constructive feedback.
This research was sponsored in part by the Patient-Centered Outcomes Research Institute Funding ME-2017C1-6413, the Army Research Office under cooperative agreements W911NF-17-1-0412, NSF Grant IIS1815674, and Ohio Supercomputer Center \cite{OhioSupercomputerCenter1987}. The views and conclusions contained herein are those of the authors and should not be interpreted as representing the official policies, either expressed or implied, of the Army Research Office or the U.S.Government. The U.S. Government is authorized to reproduce and distribute reprints for Government purposes notwithstanding any copyright notice herein.


\appendix

\newpage
\section{Appendices}
\label{sec:appendix}

\subsection{Implementation Details.}
We implemented our model in Pytorch~\citep{paszke2017automatic} and optimized it by the Adam optimizer~\citep{kingma2014adam}. The dimension of term/node embeddings is set at 128. The number of negative triples for the relational learning is set at 100. The number of association contexts to use for assumption formation is 32.  Early stopping is used when the performance in the dev set does not increase continuously for 10 epochs. {We augment the relation triples for optimizing $\mathcal{L}_r$ (Eqn.~\ref{eqn:loss_nllloss_rela}) by adding their reverse relations for better training.} We obtain DeepWalk and LINE (2nd) embeddings by OpenNE\footnote{\url{https://github.com/thunlp/OpenNE}} and word2vec embeddings by doing SVD decomposition over the shifted PPMI co-occurrence matrix~\cite{levy2014neural}. Code, datasets, and more implementation details are available online\footnote{\url{https://github.com/zhenwang9102/X-MedRELA}}.

\subsection{Training Algorithm}
\begin{algorithm}[htbp!]
\caption{CogStage Training Algorithm}
\label{alg:framework-training}
\begin{algorithmic}[1]
\INPUT Corpus Statistics $\mathcal{G}$, Gold Triples $\mathcal{P}$, Binary Relation Data $\{(h_k, t_k), y_k\}_{k=1}^{M}$

\OUTPUT {Model parameters}

\Repeat
    \State {Sample} $\{e_i\}_{i=1}^{b_1}$ with gold contexts from $\mathcal{G}$
    \For{$i\gets 1: b_1$}
        \State Calculate $p(e_j|e_i)$ and $\hat{p}(e_j|e_i)$
        \State Optimize $\mathcal{L}_n$ by Eqn.~\ref{eqn:loss_context_pred}
    \EndFor
    
    \State {Sample} $\{(h_i, r_i, t_i)\}_{i=1}^{b_2}$ from $\mathcal{P}$
    \For{$i\gets 1: b_2$}
        \State Generate $N_n$ corrupted triples
        \State Optimize $\mathcal{L}_r$ by Eqn.~\ref{eqn:loss_nllloss_rela}
    \EndFor
    
    \State {Sample} $\{(h_i, t_i), y_i\}_{i=1}^{b_3}$
    \For{$i\gets 1: b_3$}
        \State Calculate $p(e_j|h_i)$ and $p(e_j|t_i)$
        \State Get contexts $\{a_{h}^m\}_{m=1}^{N_c}$ and $\{a_{t}^n\}_{n=1}^{N_c}$
        \State Optimize $\mathcal{L}_p$ by Eqn.~\ref{eqn:loss_bce}
    \EndFor

\Until{Convergence}
\end{algorithmic}
\end{algorithm}

\begin{figure*}[!t]
    \centering
    \includegraphics[width=1\textwidth]{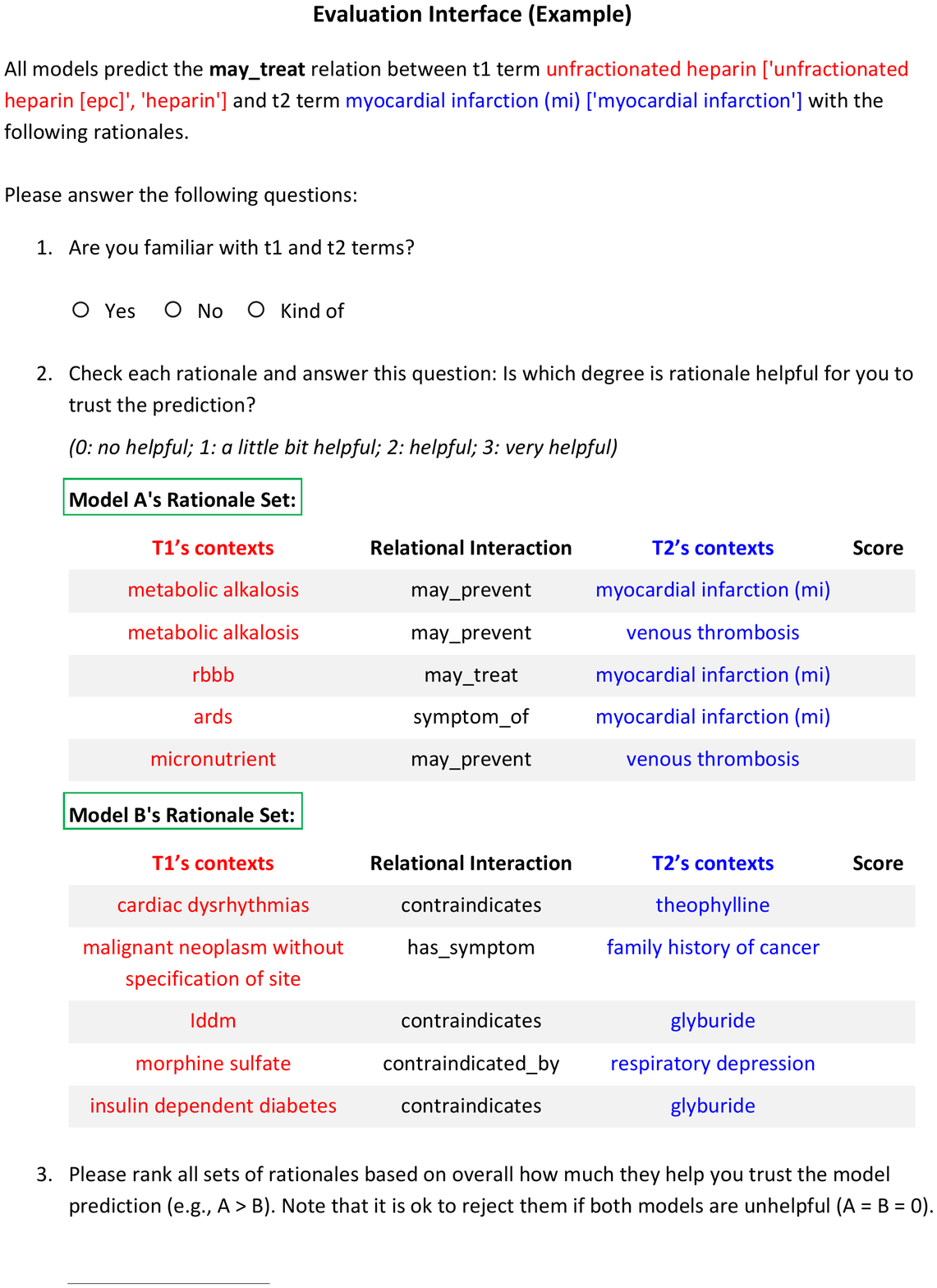}
    \vspace{-15pt}
    \caption{Evaluation interface for expert evaluation.}
    \label{fig:eval_interface}
    \vspace{-18pt}
\end{figure*}

\begin{table*}[!htbp]
\centering
\begin{tabular}{@{}lp{11cm}@{}}
\toprule
Relations & UMLS Relations \\ \midrule
May\_treat & may\_treat \\ \midrule
May\_prevent & may\_prevent \\ \midrule
Contraindicates & has\_contraindicated\_drug \\ \midrule
Causes & cause\_of; induces; causative\_agent\_of \\ \midrule
Symptom of & {\shortstack[c]{\begin{tabular}[c]{@{}c@{}}disease\_has\_finding; disease\_may\_have\_finding; has\_associated\_finding; \\ has\_manifestation; associated\_condition\_of; defining\_characteristic\_of\end{tabular}}} \\ \bottomrule
\end{tabular}
\caption{Relations {in our dataset} and their mapped UMLS semantic relations. (UMLS relation ``Treats'' does not exist in our dataset {and hence is not mapped with the ``May\_treat'' relation.})}
\end{table*}

\end{document}